\documentclass{article}
\usepackage[utf8]{inputenc}
\usepackage{svg}
\usepackage{subfig}
\usepackage[margin=2.5cm]{geometry}
\usepackage{url}
\usepackage{amssymb}

\newcommand{\projectname}{\textsc{OpenSocInt}}

\begin{document}

\title{\projectname: A Multi-modal Training Environment\\ for Human-Aware Social Navigation}

\author{Victor Sanchez $^1$, Chris Reinke$^2$, Ahamed Mohamed$^2$, Xavier Alameda-Pineda$^2$\vspace{2mm}\\
{\small E-mails: $^1$\texttt{victor.sanchez.pro@protonmail.com}, $^2$\texttt{name.lastname@inria.fr}}}
\date{Inria at Univ. Grenoble Alpes, LJK, CNRS}

% \email{victor.sanchez.pro@protonmail.com}
% \affiliation{%
%   \institution{Inria at Univ. Grenoble Alpes}
%   \city{Grenoble}

% \author{}
% \email{chris.reinke@inria.fr}
% \affiliation{%
%   \institution{Inria at Univ. Grenoble Alpes}
%   \city{Grenoble}

% \author{}
% \email{ahamed.mohamed@inria.fr}
% \affiliation{%
%   \institution{Inria at Univ. Grenoble Alpes}
%   \city{Grenoble}

% \author{}
% \email{xavier.alameda-pineda@inria.fr}
% \affiliation{%
%   \institution{Inria at Univ. Grenoble Alpes}
%   \city{Grenoble}
% \renewcommand{\shortauthors}{Sanchez et al.}


\maketitle

\begin{abstract}
    In this paper, we introduce \projectname, an open-source software package providing a simulator for multi-modal social interactions and a modular architecture to train social agents. We described the software package and showcased its interest via an experimental protocol based on the task of social navigation. Our framework allows for exploring the use of different perceptual features, their encoding and fusion, as well as the use of different agents. The software is already publicly available under GPL at \url{https://gitlab.inria.fr/robotlearn/OpenSocInt/}.
\end{abstract}

\section{Introduction}
\begin{figure}[h!]
    \centering
    \includegraphics[width=0.3\columnwidth]{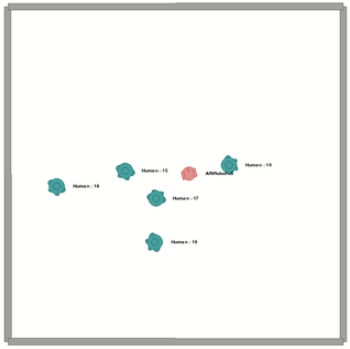}\hspace{2mm}%
    \includegraphics[width=0.3\columnwidth]{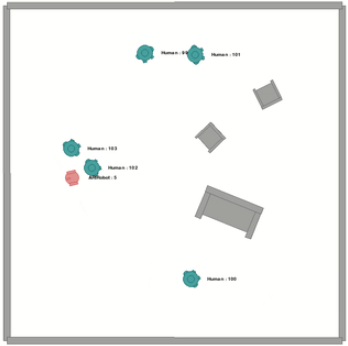}
    \caption{Two examples of the environment simulation with \projectname: (left) without and (right) with obstacles.}
    \label{fig:examples}
\end{figure}

% Human-aware social navigation \chris{the platform allows in principle for other interactions at it can simulate, gaze and  speech too. Social navigation could be shown as a first problem for which it was used, and the platform named OpenSocInt for interaction} 
The computational study of social interactions is an important field of research with many applications such as video games, virtual reality, and social robotics. The field is inherently multi-modal as, in practice, several sensors are available to describe the scene around the agent(s). Examples of such sensors are cameras, LIDARs, ultrasound, and microphones. \projectname\ provides a framework for simulating social situations and various sensor percepts, with the main aim of training agents for social interaction. The prominent example is human-aware social navigation, and two examples of this environment (with or without obstacles/furniture) are provided in Figure~\ref{fig:examples}.
% \chris{This is a bit problamatic, as the simulator can not produce RGB images from the ego view of the robot, only from the map itself}, LIDARs, and ultrasound. \projectname\ provides a framework for simulating social situations and various sensor percepts, with the main aim of training agents for social navigation. Two examples 

\section{Modular Architecture}

\begin{figure}[h!]
    \centering
    \includegraphics[width=0.75\columnwidth]{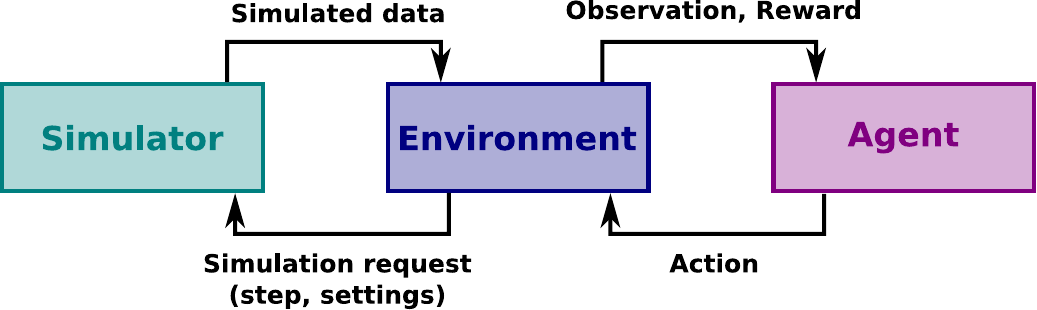}
    \caption{\projectname's modular architecture: the agent takes actions and gets observations and rewards from the environment. Following these actions, the environment queries the simulator, and computes the reward and the observations from the simulated data. }
    \label{fig:architecture}
\end{figure}

\projectname\ is structured in three different modules, namely: the simulator, the agent, and the environment, see Figure~\ref{fig:architecture}. The simulator keeps track of the current status of the scene being simulated. The agent takes actions and receives the associated rewards and the new observations that will be used to take further action. The environment acts as a mediator between the agent and the environment by communication to the environment the actions taken, and extracting observations from the new environment state.

\subsection{The Simulator Module}

This module is responsible for keeping track of the state of the scene, updating it with the information about the agent provided by the environment, and providing to the environment the updated state. Internally, the simulator knows the position of the agent, of the (dynamic and static) obstacles, see Figure~\ref{fig:examples}, and the position (angle and distance) of the agent's goal relative to the agent.

\begin{figure}[t]
    \centering
    \includegraphics[height=5cm]{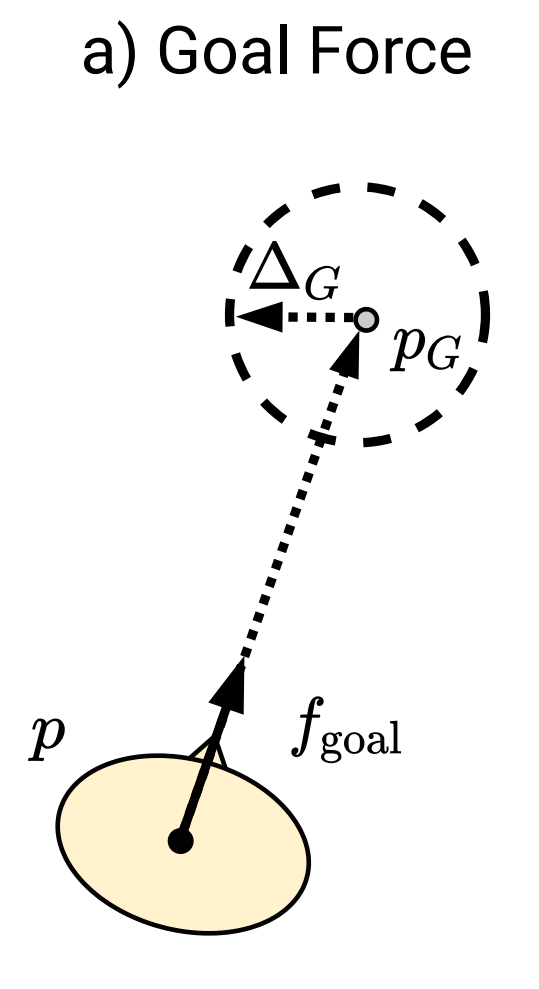}
    \hspace{4mm}
    \includegraphics[height=5cm]{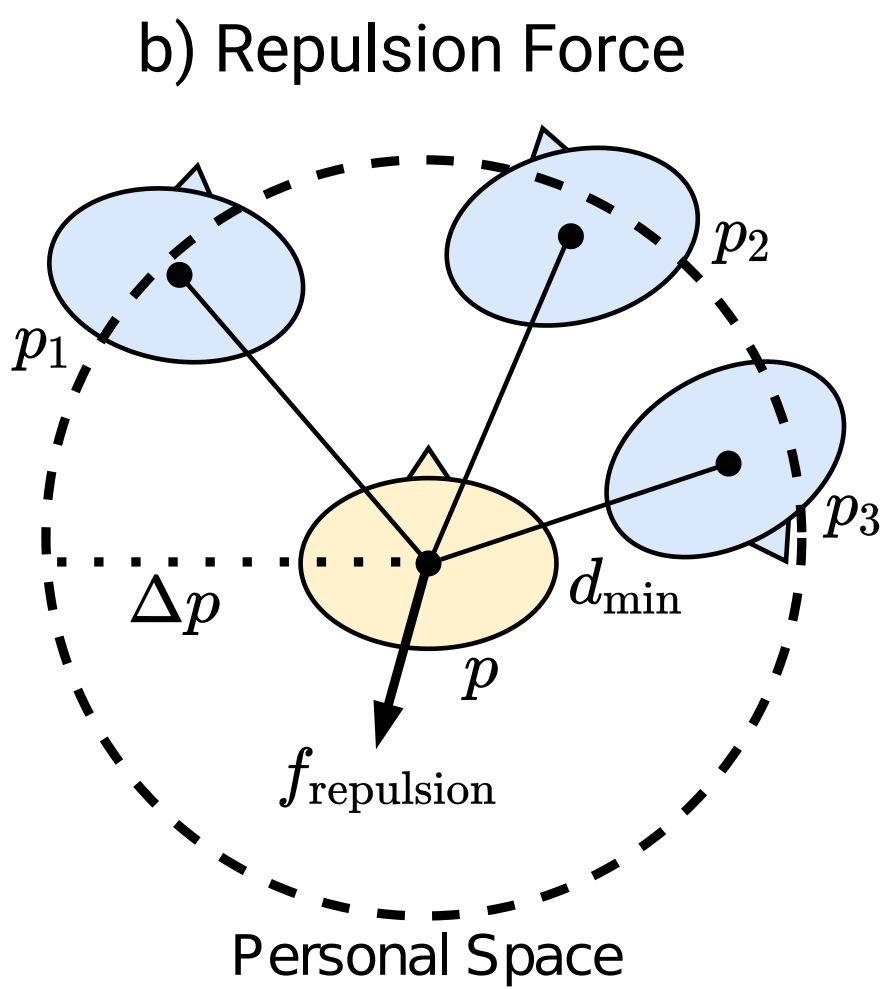}
    \caption{Diagram of the goal force (left) and social/repulsion force (right) used in the simulator module to update the state of the simulation.\label{fig:forces}}
\end{figure}

In order to compute the updated state, the simulator needs to compute the next position of the agent (given the action) and the next position of the dynamic obstacles (persons). We hypothesise that each person has a goal, and the next position of each person will be computed using a weighted sum of a \textit{social force model}~\cite{pedica2008social} and a \textit{goal force model}. While for the later, the force is oriented toward the goal and proportional to the distance (with maximum unitary value), the later is based on a repulsion force to avoid humans to be too close to each other, see Figure~\ref{fig:forces}. This repulsion force does not take the agent into account as preliminary experiments showed that the agent will learn to exploit this repulsion to push humans away and go directly to its goal.

\subsection{The Environment Module}

This module aims to facilitate the communication between the agent and the simulator. On the one hand, the actions taken by the agent are sent to the environment and then the simulator so as to step onto the simulation time. In the simulator, the internal state will be updated according to the action received, and the new state will be sent to the environment module, which will then translate the new state into one or several sensing modalities.

The first sensing modality encloses the coordinates (either in Cartesian or Polar format) of the nearest obstacle (whether it is static or dynamic). This sensing modality has the advantage of being very simple, and the disadvantage of missing a lot of information. A second approach is to convert the state onto LIDAR data around the robot, in the form of a Raycast. The data dimensionality is fixed, and depends on the desired accuracy. For example, a distance measurement at every degree around the robot implies a vector of 360 values. The format is slightly more complex than the previous one, but encodes much more information. Occluded obstacles, however, are not represented when using LIDAR. A third approach is to use a local egocentric occupancy grid (LEOG) that represents the occupancy of the space around a robot. Each element of the grid is either one if that element is occupied and zero otherwise. The size of this representation depends on the map size and resolution. For example, a $6~\textrm{m}\times 6$~m grid with a precision of $0.1$~m per pixel yields $60\times 60$ image. The environment can convert the simulation state into one or more of these representations depending on the input taken by the agent. Other representations can be implemented and added to the package very easily.

In addition to the representation of the simulation state, the environment also forwards the relative position of the agent's goal and the reward computed using the simulator's data.

\begin{figure}[t]
    \centering
    \includegraphics[width=0.4\columnwidth]{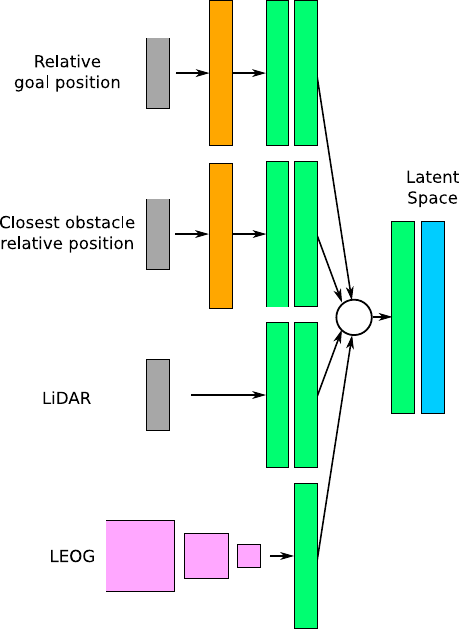}
    \caption{Architectures for processing the various input modalities. Convolutional layers (for LEOG) are in pink, fully connected layers in green, and radial basis fuction layers in orange. The circle stands for the concatenation-based multi-modal fusion. The various implemented agents always build from the latent space, shown in blue.}
    \label{fig:modality-architecture}
\end{figure}

\subsection{The Agent Module}
The agent receives the observations (representation of the simulation state, as one or multiple modalities), the relative goal's position, and the obtained reward.

At training time, the observations are forwarded through the network, that outputs the action taken, as well as the other values needed for training. These values will depend on the deep RL paradigm being used (see below). The loss is computed and back-propagated through the network. The action chosen is then forwarded to the environment, thus closing the loop in Figure~\ref{fig:architecture}.

The multi-modal agent architecture inputting the various modalities is shown in Figure~\ref{fig:modality-architecture}. A latent space (blue box) is learn from one or more modalities plus the agent's goal. The local egocentric occupancy grid (LEOG) is processed by a few convolutional layers (in pink), then flattened and transformed via one or more fully connected layers (in green). The LIDAR (RayCast) and the other two modalities are transformed via one or more fully connected layers. In order to facilitate learning from very low-dimensional inputs (e.g.\ the relative position of the goal and closest obstacle) a radial basis function~\cite{montazer2018radial} layer (in orange) is used to increase the representation's dimensionality before the fully connected layers.
The one or multiple modalities are concatenated into the latent representation, that is then used by the various reinforcement learning agents implemented.

All the implemented agents exploit the same latent representation (blue in the image) and cover the following deep RL paradigms: twin-delayed deep deterministic policy gradient (TD3)~\cite{dankwa2019twin}, deep deterministic policy gradient (DDPG)~\cite{lillicrap2015continuous}, advantage actor-critic (A2C)~\cite{kumar2023sample} and soft-actor critic (SAC)~\cite{haarnoja2018soft}. 

\subsection{Repository}
The code repository is publicly available,\footnote{\url{https://gitlab.inria.fr/robotlearn/OpenSocInt/}}, and has the corresponding \texttt{agent} and \texttt{env} folders, as well as a few other utility folders for creating datasets and training recipies. The simulator is an automatically linked external module that is kept in a different repository so that users interested only in the simulator do not need to download the entire \projectname\ package. A detailed explanation of the scripts, options, and visualisation tools can be found in the \texttt{README.md} file.

\section{Experimental Protocol}
In order to validate the use of the proposed software package, we have conducted a series of experiments training reinforcement learning agents via the environment and simulator described above. We describe here the key information of our experimental protocol.
% \textbf{Pour Xavi : La figure qui compare les agents est prête et elle est chargée dans le dossier image.}

\subsection{Reward functions}
During training, the agent has to learn the optimal behavior. To that aim, the agent interacts with the environment by taking actions during a given number of episodes. An episode terminates under on of the following three conditions: reaching the goal, colliding with an object/human, or the agent reaches the maximum number of steps without collision or  reaching the goal (the episode state is ``Truncated''). The agent’s optimal behaviour is reached by maximising the cumulative reward. The reward function in intermediate steps (before reaching the goal/colliding) writes as follows:
\begin{equation}
    r_{\textsc{int}}(t) = r_{\textsc{step}}(t) + r_{\textsc{goal-d}}(t) + r_{\textsc{social}}(t),
\end{equation}
where $r_{\textsc{step}}(t) =-\omega_{step}$ penalises taking unnecessary steps, and $r_{\textsc{goal-d}}$ and $r_{\textsc{social}}$ correspond to the goal direction and social repulsion forces shown in Figure~\ref{fig:forces}, multiplied by coefficients $\omega_{\textsc{goal-d}}$ and $\omega_{\textsc{social}}$, respectively. When reaching the goal/colliding (end of the episode), the reward writes as follows:
\begin{equation}
    r_{\textsc{end}}(t) = \omega_{\textsc{goal-r}}.\gamma_{\textsc{goal-r}}(t) + \omega_{\textsc{coll}}.\gamma_{\textsc{coll}}(t),
\end{equation}
where $\gamma_{\textsc{goal-r}}(t)=1$ if the goal is reached and 0 otherwise and $\gamma_{\textsc{coll}}(t)$ is defined analogously for the collision. The values for the weigthing coefficients can of course be parametrised in our package. In the experiments reported below, we use the following numbers:  $\omega_{\textsc{goal-r}}=500$, $\omega_{\textsc{coll}}=-500$,  $\omega_{step}=-5$, $\omega_{\textsc{goal-d}}=10$, and $\omega_{\textsc{social}}=-100$.

% the number of steps for joining the goal position. To give the agent an idea of improvement when going towards the goal position we provide a reward that is the difference between the former distance to the goal and the actual distance to the goal: $r_{goal\_direction}(t) = \omega_{goal\_direction}.[d_{(agent/goal)}(t-1) - d{(agent/goal)}(t)]$. So as to give the agent a social behavior we provide a social score that is taken from previous work \cite{truong_approach_2017}  $r_{social}(t) =−\omega_{social}.\theta_{social}(t)$.

% and the reward signal (or reward function) specifies what the agent must achieve. Depending on the behavior of the agent, the reward can be \textit{final} meaning that the episode terminates or \textit{non-final} otherwise. The final reward is defined as : 
% \begin{equation*}
%     r_{final}(t) = \omega_{goal\_reached}.\gamma_{goal\_reached}(t) + \omega_{collision}.\gamma_{collision}(t)
% \end{equation*}
% where $\gamma_{goal\_reached}=1$ if the goal is reached and 0 otherwise and $\gamma_{collision}=1$ if there is a collision (human or object related) and 0 otherwise. \\
% We then defined the non-final reward as : 

% \begin{table}[]
% \begin{tabular}{|c|c|}
% \hline
%  $\omega_{goal\_reached}$  &  500 \\ \hline
%  $\omega_{collision}$  &  -500  \\ \hline
%  $\omega_t$  &  -5  \\ \hline
%   $\omega_{goal\_direction}$  &  10  \\ \hline
%   $\omega_{social}$  &  -100  \\ \hline
% \end{tabular}
% \caption{Weigth values for each reward component.}
% \end{table}

\subsection{Metrics}
We report the evolution of three metrics along the training. First, the percentage of episodes that end with the agent reaching the goal. Second, the percentage of episodes that end with the agent colliding either with a static or with a dynamic obstacle. Third, the percentage of episodes that run until the maximum number of steps without reaching the goal or colliding, see the previous section. Importantly, these metrics are computed in a set of episodes that does not belong to the training set, and is therefore never part of the replay buffer. More precisely, every 50 training episodes, the model is evaluated on 20 test episodes, until reaching 700 training episodes (or 280 test episodes). This is done 5 times per configuration. The mean and standard deviation of each metric over a sliding window of 10 test episodes is the reported.

\section{Results}

\subsection{The impact of the LEOG encoding}
\begin{figure}
    \centering
    % \subfloat[first caption.\label{fig:1a}]{\includegraphics[width=0.2\textwidth]{fig1a}}\hfill
    \subfloat[LEOG encoder.\label{fig:encoders}]{\includegraphics[width=.24\textwidth]{images/leog_input_comparison_with_truncated.png}}
    \subfloat[Features.\label{fig:features}]{\includegraphics[width=.24\textwidth]{images/input_comparison_with_truncated.png}}
    \subfloat[Feature fusion.\label{fig:fusion}]{\includegraphics[width=.24\textwidth]{images/raycast_input_comparison_with_truncated.png}}
    \subfloat[RL paradigms.\label{fig:agents}]{\includegraphics[width=.24\textwidth]{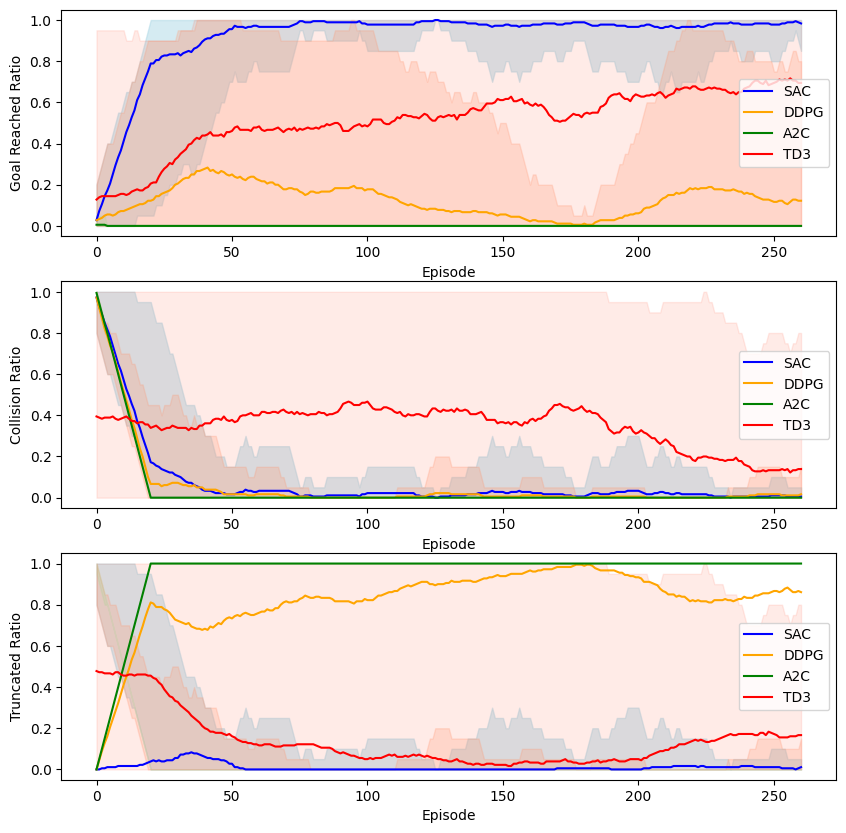}}
    \caption{Mean and standard deviation of the success, collision, and truncated rates of the different settings evaluated. (a) Different LEOG encoding strategies: pre-trained frozen, pre-trained and fine-tuned, trained from scratch. (b) Different features: LEOG, RayCast and Closest Object. (c) Feature fusion: LEOG, RayCast, and LEOG+RayCast. (d) Various RL paradigms: SAC, TD3, DDPG, and A2C.}
\end{figure}

While the RayCast and the closest obstacle feature are relatively compact, LEOG are raw features that can be compressed. To this aim, we proposed to encode them via a convolutional neural network. However, the importance of the LEOG encoding pre-training has to be assessed.
In order to pre-train the encoder, we simply let the agent wonder around the environment to collect data using a random policy. If \projectname\ is to be interfaced with a custom simulator that is computationally to heavy to pre-train an encoder on-line, our code provides also the option of collecting a dataset in advance/use an existing dataset to pre-train the encoder. Obviously key training parameters (e.g.\ batch size) can be parametrised, and we also provide tools for visualising the training progress. 
% \textcolor{green}{The encoder can be trained using two methods, first by directly using the simulator to obtain observations during training, second by gathering a datasets of occupancy grids and training offline. Both these methods are available in the code provided. 
% Each training can be configured to use 2 types of environnements, a default configuration is present in the example training code also provided with the code. After choosing key parameters such as batch size, number of episodes or epochs (based of the type of training), learning rate, the training can be started. The checkpoints generated during the training can be used to visualize the progress of the training with the plotting code}. 
In Figure~\ref{fig:encoders}, we report both metrics in three different situations, namely: a pre-trained encoder, a pre-trained encoder that is fine-tuned when training the RL agent, an untrained encoder that is trained from scratch together with the RL agent. 

The results clearly indicate that, while all three LEOG encoder training strategies converge to an excellent performance, there are difference in terms of standard deviation during training. Indeed, learning the LEOG encoder from scratch at the same time as the agent, results in higher standard deviations during training time. In addition, we also observe the natural superiority of the pre-trained and frozen LEOG encoder at the beginning of the agent training: the inductive bias of the encoder pre-training helps considerably.

% \begin{figure}
%     \centering
%     \includegraphics[width=\columnwidth]{images/leog_input_comparison_with_truncated.png}
%     \caption{Success and collision rates of the different pre-training and training strategies of the LEOG encoder using the soft-actor critic agent: \textcolor{red}{Mention colors and modalities.}}
%     \label{fig:encoders}
% \end{figure}

\subsection{The impact of the modalities}
We first analyse the impact of the features used as network input. As a reminder, there are three options: the polar coordinates relative to the agent of the closest obstacle, the LIDAR or RayCast information, and the local egocentric occupancy grid (LEOG). Figure~\ref{fig:features}, reports the metrics when inputting these modalities. 

We can observe that, while asymptotically old modalities converge to a similar performance, there are notable differences specially at the first stages of training. Indeed, the superiority of simpler features such as RayCast and Closest Object is natural given that they are easier to exploit. However, the LEOG features provide the same mean performance with small standard deviation (thus more reliability) when trained with more data. 

% \textcolor{red}{Discuss the differences}.

% \begin{figure}
%     \centering
%     \includegraphics[width=\columnwidth]{images/input_comparison_with_truncated.png}
%     \caption{Success and collision rates when input the three different modalities to the soft-actor critic agent: \textcolor{red}{Mention colors and modalities.}}
%     \label{fig:features}
% \end{figure}

\subsection{The impact of fusing modalities}
In addition of comparing the various features in the previous section, we have also considered the interest of merging different feature modalities, so as to evaluate the interest of multi-modal agents. To this aim, Figure~\ref{fig:fusion} presents the results comparing the fusion of RayCast and LEOG modalities to having the two modalities independently.

While at a first glance one could think that the fusion does not bring a strong improvement, a more detailed analysis points to the opposite direction. First of all, during the first half of the training, the fusion of the two modalities exhibits better average performance than both modalities independently. In addition, the method exploiting the two modalities has considerably small standard deviation than the mono-modal models. Specifically, we observe sporadic increases of truncated episodes for both mono-modal methods at different points of the training, while this phenomenon is not observed in the multi-modal case. Overall, the main benefit of the mutli-modal strategy seems to be a more effective learning in the low-data regime, and a more stable training.

% \begin{figure}
%     \centering
%     \includegraphics[width=\columnwidth]{images/raycast_input_comparison_with_truncated.png}
%     \caption{Success and collision rates when input two modalities and their fusion to the soft-actor critic agent: \textcolor{red}{Mention colors and modalities.}}
%     \label{fig:fusion}
% \end{figure}

\subsection{The impact of the RL paradigm}
As mentioned above, the \projectname\ software package is a simulation environment designed for reinforcement learning. In this regard, we have implemented and compared several reinforcement learning paradigms when input the same modality (LEOG). Figure~\ref{fig:agents} reports the results of the four RL paradigms evaluated, namely SAC, TD3, DDPG, and A2C. What is most interesting with these last results is that the different RL paradigms exhibit very different behaviours. First, A2C could be considered as learning a ``people-avoiding'' policy, regardless of the goal. Indeed, the truncated ratio grows to 1, while the goal reaching and collision quickly converge to 0. Second, the DDPG exhibits a similar learned behaviour, with the difference that in some situations it manages to reach the goal. Next inline would be TD3, that exhibits an almost random behaviour, with many episodes stopped with a collision or truncated, while the majority by reaching the goal. Finally, SAC exhibits the best behaviour, rapidly decreasing the number of collisions with almost no truncated episodes, thus reaching the goal in the most cases since the first training stages.

\section{Conclusions}
In this paper we introduce \projectname, an open-source software package dedicated to training RL-based autonomous agents for multi-modal social interaction. We described its modular architecture and showcase its interest by conducting and reporting a series of experiments in the social navigation application. Different agents and features (as well as their encoding and fusion), are evaluating, thus putting forward the versatility and interest of \projectname. Future development directions will consist in enriching the simulator with social cues, and training agents for other tasks.

\section*{Acknowledgements}
We would like to warmly thank Alex Auternaud, Ga\"etan Lepage, and Anand Ballou for the valuable discussions and feedback provided during the development and evaluation of the \projectname\ software package.

\bibliographystyle{plain}
\bibliography{refs}

\end{document}